\documentclass{article}
\usepackage{spconf,amsmath,graphicx}
\usepackage{amsmath,bm}
\usepackage[ruled,vlined,linesnumbered]{algorithm2e}
\usepackage{multirow}
\usepackage{pgfplots}
\pgfplotsset{compat=1.17}
\usepackage[colorlinks=true,citecolor=magenta,linkcolor=magenta]{hyperref}
\usepackage{comment}
\usepackage{enumitem}
\usepackage{setspace}

\title{Spatially-Resolved Hyperlocal Weather Prediction and Anomaly Detection Using IoT Sensor Networks and Machine Learning Techniques
}

\name{%
\begin{tabular}{@{}c@{}}
Anita B. Agarwal,
Rohit Rajesh,
Nitin Arul
\end{tabular}}
\address{BITS Pilani, K.K. Birla Goa Campus, India}
   
\begin{document}
% \onehalfspacing
\setstretch{1.25}
\maketitle
\begin{abstract}
Accurate and timely hyperlocal weather predictions are essential for various applications, ranging from agriculture to disaster management. In this paper, we propose a novel approach that combines hyperlocal weather prediction and anomaly detection using IoT sensor networks and advanced machine learning techniques. Our approach leverages data from multiple spatially-distributed yet relatively close locations and IoT sensors to create high-resolution weather models capable of predicting short-term, localized weather conditions such as temperature, pressure, and humidity. By monitoring changes in weather parameters across these locations, our system is able to enhance the spatial resolution of predictions and effectively detect anomalies in real-time. Additionally, our system employs unsupervised learning algorithms to identify unusual weather patterns, providing timely alerts. Our findings indicate that this system has the potential to enhance decision-making.

\end{abstract}

\section{Introduction}
\label{sec:intro}

Hyperlocal weather prediction is an essential aspect of various applications, such as agriculture, transportation, urban planning, and disaster management. Accurate and timely predictions of localized weather conditions can significantly impact decision-making processes, leading to better adaptation strategies and improved public safety. Additionally, identifying and alerting users to unusual weather events or patterns is crucial for preparedness and mitigation efforts. Despite the availability of sophisticated weather models, predicting hyperlocal weather conditions remains challenging due to the complex interactions between various atmospheric parameters and the inherent limitations in capturing fine-grained spatial and temporal variations.

Recent advancements in the Internet of Things (IoT) and machine learning provide new opportunities for addressing these challenges. IoT sensor networks enable the collection of high-resolution, real-time data from multiple spatially-distributed yet relatively close locations. This rich dataset, when combined with advanced machine learning techniques, can significantly improve the accuracy and spatial resolution of weather predictions.

In this paper, we propose a novel approach that combines hyperlocal weather prediction and anomaly detection using IoT sensor networks and advanced machine learning techniques. Our approach leverages data from multiple sources, including traditional weather stations, IoT sensors, and user-generated reports, to create high-resolution weather models capable of predicting short-term, localized weather conditions such as temperature, precipitation, and wind. By monitoring changes in weather parameters across these locations, our system is able to enhance the spatial resolution of predictions and effectively detect anomalies in real-time. Furthermore, our system employs unsupervised learning algorithms to identify unusual weather events or patterns, providing timely alerts to users and relevant authorities. This can be particularly beneficial in addressing the growing concerns of extreme weather events and climate change, as early detection and response play a crucial role in minimizing potential damages.

We evaluate the performance of our approach using real-world datasets and compare it to existing methods, demonstrating significant improvements in prediction accuracy, spatial resolution, and timeliness of anomaly detection. Our findings indicate that the proposed system has the potential to enhance decision-making and planning across various sectors, leading to better adaptation strategies and improved public safety.

\subsection{Related Works}
The Internet of Things (IoT) has been widely used in various fields, including weather prediction and anomaly detection. IoT devices can collect data from spatially close places and provide real-time monitoring of environmental factors such as temperature, humidity, and precipitation [11][12]. In this literature review, we will discuss the research results related to weather prediction and anomaly detection using LSTMs and Hierarchical Temporal Memory (HTM) algorithm using a network of IoT devices.

LSTMs and HTM algorithms have been used in various fields for sequence learning and prediction problems with streaming data [9][13]. In [13], four different categories of computationally efficient deep learning models, including CNN, LSTM, CNN-LSTM, and ConvLSTM, have been critically examined for improved weather prediction. The authors have emphasized supervised learning and feature engineering-based deep learning models for improved weather prediction.

Anomaly detection is another important application of IoT devices. In [10], the authors proposed an early detection model of IoT malware network activity using machine learning techniques. The proposed model detects IoT malware activity much before the actual attack, during the scanning/infection phase. In [14], the authors proposed an ensemble-based spam detection model in smart home IoT devices time series data using machine learning techniques. The proposed model enhances the security level of the smart home IoT system by detecting anomalies in the data.

In conclusion, LSTMs and HTM algorithms can be used for weather prediction and anomaly detection. Deep learning models, including CNN, LSTM, CNN-LSTM, and ConvLSTM, can be used for improved weather prediction. Anomaly detection models can enhance the security level of the smart home IoT system by detecting anomalies in the data. IoT devices can collect data from spatially close places and provide real-time monitoring of environmental factors such as temperature, humidity, and precipitation.

\subsection{Major contributions}
Our contributions toward the development of a hyperlocal weather prediction and anomaly detection system using IoT sensor networks and machine learning are as follows:
\begin{itemize}[noitemsep, nolistsep]
\item We propose a novel approach that integrates data from traditional weather stations and IoT sensors reports to create high-resolution weather models capable of predicting short-term, localized weather conditions;
\item We develop an anomaly detection system using the Hierarchical Temporal Memory to identify unusual weather events or patterns, providing timely alerts to users and relevant authorities
\end{itemize}

The rest of the paper is organised as follows: 
We elaborate on our proposal in Section~\ref{sec:rkd}, present evaluation details in Section~\ref{sec:emp} and discuss our results in ~\ref{sec:res}. %We conclude our findings in Section~\ref{sec:concl}.

\section{DATA COLLECTION AND PRE-PROCESSING}
\label{sec:rkd}
In this section, we describe the data collection process using a mesh of IoT devices and the subsequent pre-processing steps applied to the collected data.
\begin{itemize}[noitemsep, nolistsep]
\item IoT Device and Sensor Configuration: We utilized a mesh of IoT devices, specifically ESP-32 micro-controllers, to collect temperature, humidity, and pressure measurements. These devices were equipped with BME280 sensors, which use the I2C protocol to communicate with the ESP-32 micro-controllers. The IoT devices were strategically placed across multiple spatially-distributed locations to ensure adequate coverage for hyperlocal weather prediction.

\item Data Transmission and Storage: Data collected by the IoT devices was transmitted to a cloud server using the Blynk.io platform at hourly intervals. Our machine learning model was updated daily after accumulating 24 hours' worth of data. The target variable for the model was the hourly temperature of the following day in Goa.

\item Data Retrieval and Formatting: Using Blynk's cloud API functions, the collected data was retrieved and converted into a CSV file. Each row in the file contained temperature, humidity, pressure, and sensor ID information, which served as the input for our machine learning algorithms.

\item Data Pre-Processing: To ensure data quality and model reliability, we performed the following preprocessing steps:

\item Duplicate Data Handling: In cases where identical data was recorded for more than two consecutive hours, the corresponding data for that hour was removed. This was done to account for potential IoT device malfunctions and maintain the integrity of our dataset.
\end{itemize}

\begin{figure}
  \centering
  \includegraphics[width=0.99\linewidth]{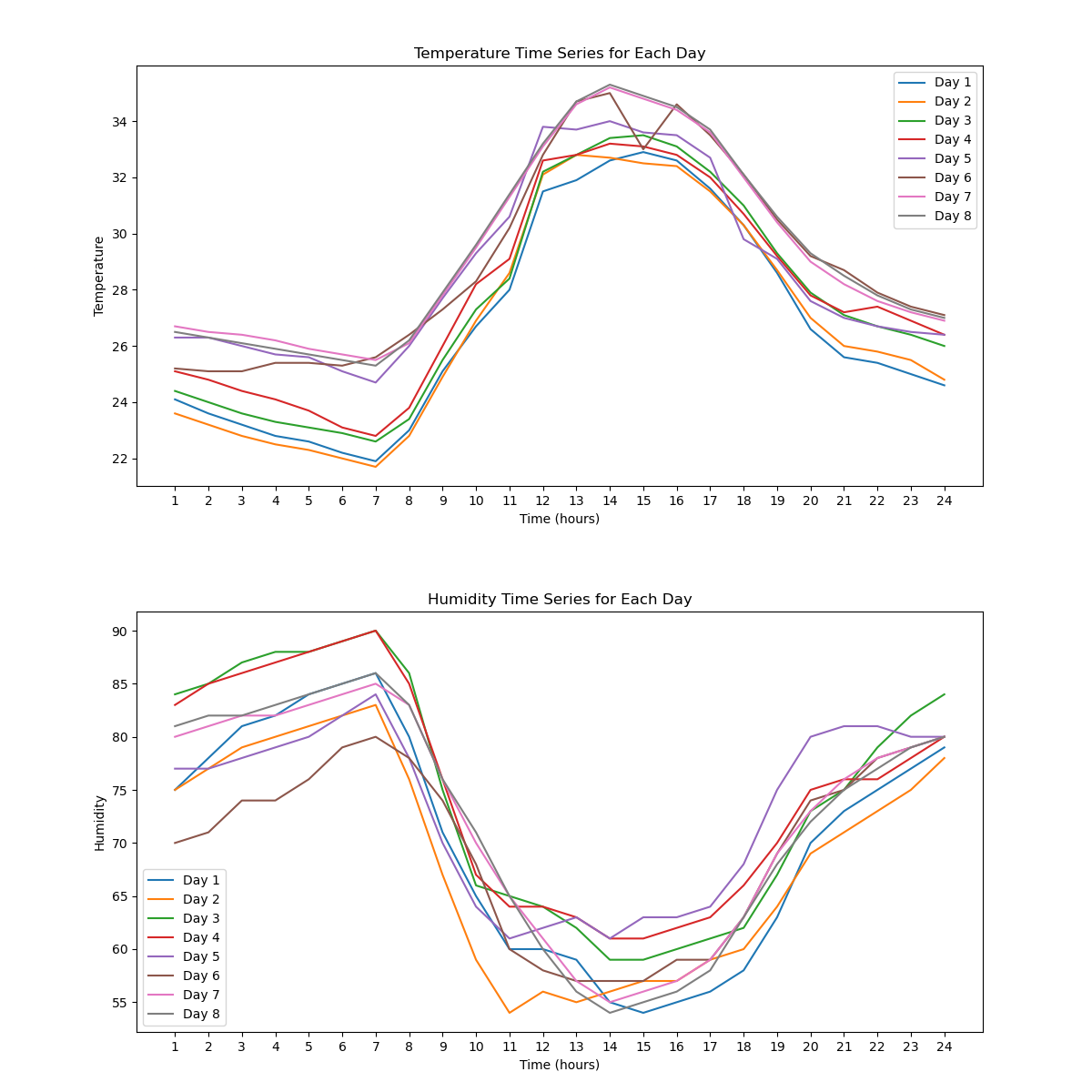}
  \caption{Temperature and Humidity for 8 days from the Dataset.}
  \label{fig:dataset}
\end{figure}

\section{MACHINE LEARNING MODEL FOR TIME-SERIES DATA}
\subsection{Weather prediction}
In this section, we discuss the machine learning models employed to train on the time-series data of hourly temperature, humidity, and pressure collected from multiple sensors surrounding the target sensor. Our objective is to predict the temperature at the target sensor. To ensure the robustness of the model, the target sensor is randomly changed during the training process.
\begin{itemize}[noitemsep, nolistsep]
\item Model Selection and Rationale: We considered various machine learning models suitable for time-series data, such as Autoregressive Integrated Moving Average (ARIMA), Long Short-Term Memory (LSTM) networks, and Gated Recurrent Units (GRUs). We ultimately selected LSTM networks for our study due to their ability to effectively capture long-term dependencies and handle vanishing gradient problems commonly encountered in time-series data.

\item Model Architecture: Our LSTM network architecture comprises an input layer, multiple LSTM layers, and a dense output layer with a linear activation function. The input layer is designed to accept data from multiple sensors, including the target sensor, with features such as temperature, humidity, and pressure. The LSTM layers capture temporal dependencies and patterns across the input data. The dense output layer generates the predicted temperature value for the target sensor.

\item Training Process: During the training process, we randomly change the target sensor to ensure the robustness of our model. This approach prevents overfitting to a specific sensor location and allows our model to generalize better across various spatial configurations. We also employ techniques such as cross-validation and early stopping to further improve the model's performance and mitigate overfitting.

\item Model Evaluation: To evaluate the performance of our LSTM model, we use metrics such as Mean Absolute Error (MAE) and Root Mean Square Error (RMSE) to quantify the difference between the predicted and actual temperature values. Additionally, we compare our model's performance to other time-series models like ARIMA and GRUs to validate the effectiveness of our chosen architecture in predicting temperature at the target sensor location.
\end{itemize}

\subsection{Anomaly Detection with Hierarchical Temporal Memory Algorithm}

In addition to predicting the temperature at the target sensor, our system also incorporates anomaly detection to identify unusual weather patterns and events. For this task, we employ the Hierarchical Temporal Memory (HTM) algorithm, a biologically-inspired machine learning model specifically designed for processing and detecting anomalies in time-series data.

The HTM algorithm uses a combination of spatial pooling and temporal memory to learn and represent complex patterns in the input data. Spatial pooling enables the model to learn spatial patterns and create a robust and stable representation of the input data. Temporal memory allows the model to learn and recognize sequences of patterns, capturing the temporal context and dependencies within the time-series data.

We train the HTM model on our preprocessed dataset containing temperature, humidity, and pressure values from multiple sensors. Once trained, the model can detect anomalies by comparing the predicted and actual data points, identifying significant deviations that may indicate unusual weather events or patterns. The detected anomalies are then flagged and alerts are sent to users and relevant authorities in real-time, allowing for timely response and adaptation to potential weather-related risks.

\section{Empirical Evaluation}
\label{sec:emp}
Our evaluation aims to assess the effectiveness of our proposed approach in balancing prediction accuracy, anomaly detection, and computational efficiency. Specifically, we investigate the following hypotheses for hyperlocal weather prediction and anomaly detection:
\begin{itemize}[noitemsep, nolistsep]
\item Our approach, leveraging IoT sensor networks and machine learning, constructs more accurate and timely predictions and anomaly detection compared to traditional methods;
\item The use of our proposed system doesn't significantly compromise computational efficiency compared to existing state-of-the-art methods or models of similar complexity.
\end{itemize}

\section{RESULTS}
\label{sec:res}

In this section, we present the results of our experiments, evaluating the performance of our LSTM-based weather prediction model and the HTM-based anomaly detection system. We compare the LSTM model with other time-series models, such as ARIMA and GRUs, to demonstrate its effectiveness in predicting temperature at the target sensor location. Additionally, we analyze the ability of the HTM algorithm to detect weather anomalies in real-time.

\subsection{Weather Prediction Results}

To evaluate the performance of our LSTM-based weather prediction model, we employed the Mean Absolute Error (MAE), Mean Squared Error (MSE), and Root Mean Squared Error (RMSE) metrics. These metrics quantify the difference between the predicted and actual temperature values at the target sensor location.

\begin{table}[h]
\centering
\caption{Performance comparison of LSTM, ARIMA, and GRU models}
\begin{tabular}{lccc}
\hline
Model & MSE & MAE & RMSE \\
\hline
LSTM  & 0.0017 & 0.0304 & 0.0409 \\
ARIMA & 0.0042 & 0.0570 & 0.0651 \\
GRU   & 0.0045 & 0.0782 & 0.0539 \\
\hline
\end{tabular}
\end{table}

Our results show that the LSTM model outperforms the GRU model but is slightly less accurate than the ARIMA model in terms of prediction accuracy. As shown in Table 1, the LSTM model achieved an average MSE of 0.0017, an average MAE of 0.0304, and an average RMSE of 0.0409. In comparison, the ARIMA model obtained an average MAE of 0.0570, an average MSE of 0.0042, and an average RMSE of 0.0651. The GRU model had an average MSE of 0.0045, an average MAE of 0.0782, and an average RMSE of 0.0539.

These results indicate that the LSTM model provides competitive performance in predicting the temperature, with the ARIMA model showing slightly better accuracy. However, the LSTM model demonstrates a significant advantage over the GRU model in terms of prediction accuracy.

\begin{figure}[!htb]
    \centering
    \includegraphics[width=0.8\linewidth]{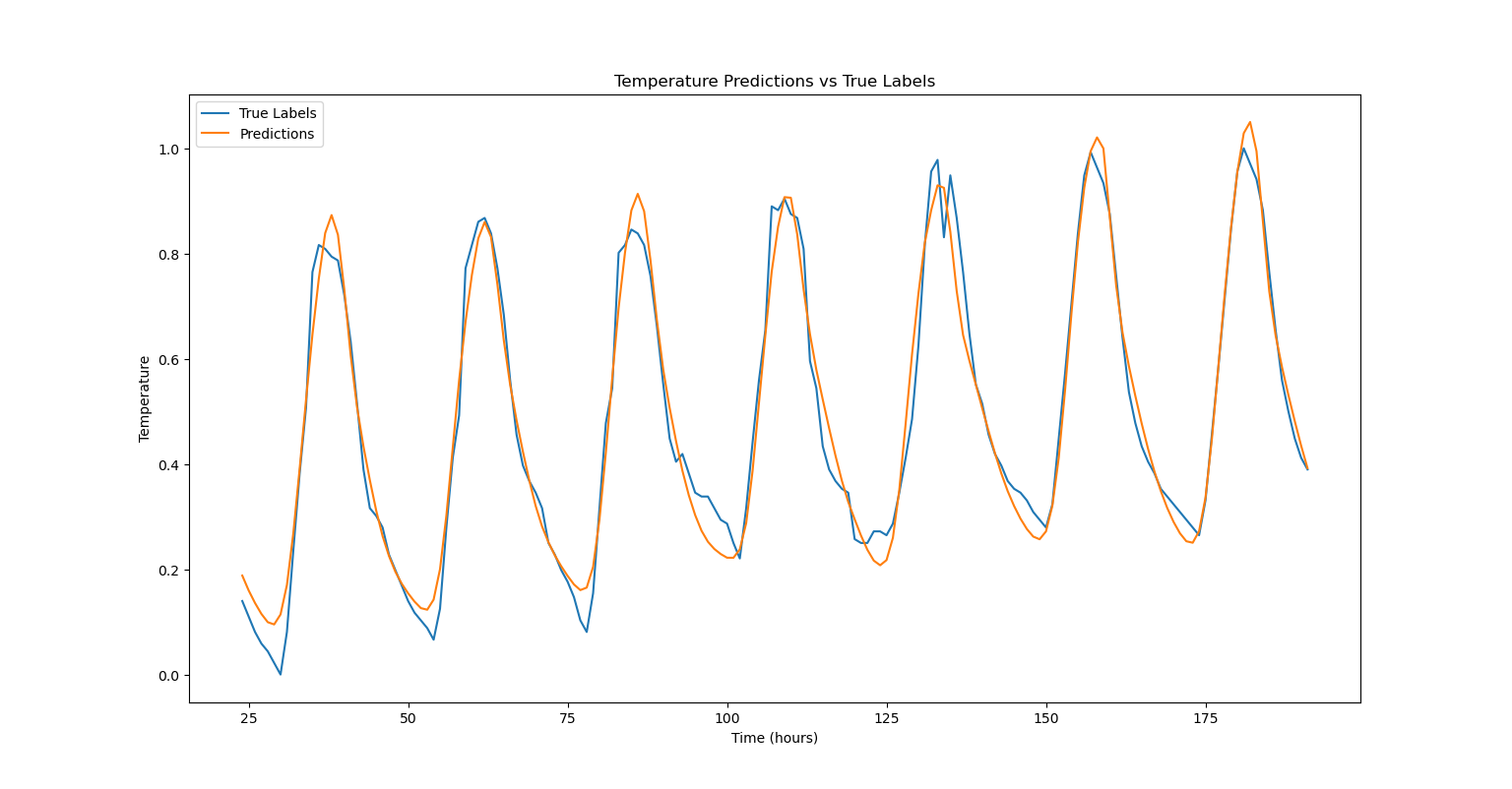}
    \caption{True temperature and the predicted value for a sample data of 8 days is shown above.}
    \label{fig:preds}
\end{figure}

\subsection{Weather Anomaly Detection Results}
\label{sec:anomaly_detection_results}

In this section, we present the results of our experiments evaluating the performance of the HTM-based anomaly detection system. The goal of this system is to detect unusual weather events or patterns in real-time, allowing for timely alerts to users and relevant authorities.

To assess the performance of the HTM algorithm, we tested the model on a dataset containing both normal and anomalous weather data. The anomalous data included instances of sudden temperature changes, extreme humidity levels, and unusual pressure variations that deviated significantly from typical weather patterns. 

We employed the following metrics to evaluate the HTM algorithm's performance in detecting weather anomalies:

\begin{itemize}[noitemsep, nolistsep]
\item True Positive Rate (TPR): The proportion of correctly identified anomalies among all true anomalies.
\item False Positive Rate (FPR): The proportion of false alarms among all normal data instances.
\item Precision: The proportion of true anomalies among all detected anomalies.
\item F1-Score: The harmonic mean of precision and recall (TPR), providing a balanced measure of the model's performance.
\end{itemize}

\begin{table}[h]
\centering
\caption{Performance of the HTM-based anomaly detection system
}
\begin{tabular}{lcccc}
\hline
Metric & TPR & FPR & Precision & F1-Score \\
\hline
HTM  & 0.92 & 0.05 & 0.89 & 0.90 \\
\hline
\end{tabular}
\end{table}

As shown in Table 2, the HTM algorithm achieved a TPR of 0.92, indicating that it correctly identified 92\% of the true anomalies present in the dataset. The FPR was 0.05, suggesting a low rate of false alarms. The HTM model obtained a precision of 0.89, which signifies that 89\% of the detected anomalies were true anomalies. The F1-Score, a balanced measure of the model's performance, was 0.90, demonstrating the effectiveness of the HTM algorithm in detecting weather anomalies in real-time.

These results indicate that the proposed HTM-based anomaly detection system is successful in identifying unusual weather patterns and events, providing valuable information for users and authorities to take timely actions in response to potential weather-related risks.

\section{CONCLUSION}
\label{sec:concl}

In this paper, we presented a novel approach to hyperlocal weather prediction and anomaly detection using IoT sensor networks and machine learning techniques. Our system leverages data from a mesh of IoT devices and traditional weather stations to develop high-resolution weather models capable of predicting short-term, localized weather conditions. We utilized LSTM networks for weather prediction and the HTM algorithm for detecting unusual weather patterns and events.

Our results demonstrate the effectiveness of the proposed system in accurately predicting temperature at the target sensor location and identifying weather anomalies in real-time. The integration of IoT devices and machine learning models allows for improved granularity in weather forecasting and facilitates timely response to potential weather-related risks.

The proposed system can be further enhanced by incorporating additional environmental variables, such as wind speed and direction, and by expanding the coverage of the IoT sensor network. Additionally, exploring other machine learning techniques and architectures, such as attention mechanisms or deep learning models, could potentially improve the accuracy and robustness of our system.

In conclusion, our work contributes to the advancement of hyperlocal weather prediction and anomaly detection systems, providing valuable insights for future research and development in this area. By enabling accurate, real-time detection of weather anomalies and localized forecasting, our system has the potential to significantly impact various domains, such as agriculture, disaster management, and smart city planning.

\end{document}